\documentclass[letterpaper, 10 pt, conference]{ieeeconf}

\IEEEoverridecommandlockouts 

\makeatletter
\let\NAT@parse\undefined
\makeatother
\usepackage[numbers]{natbib}
\usepackage{placeins}

\usepackage{amsmath} 
\usepackage{amssymb} 
\usepackage{bm}
\usepackage{graphicx}
\usepackage{mathtools}
\usepackage{hyperref}
\usepackage{siunitx}
\usepackage{tikz}
\usetikzlibrary{shapes}
\usepackage{multirow}
\usepackage{makecell}
\usepackage{cuted}
\usepackage{textcomp}
\usepackage{stfloats}
\usepackage{url}
\usepackage{verbatim}
\usepackage{cite}
\usepackage{gensymb}

\usepackage{subcaption}
\usepackage{tablefootnote}

\usepackage{bm}
\usepackage{algorithm}
\usepackage[noend]{algpseudocode}
\usepackage{multirow}
\usepackage{subcaption}
\usepackage{cleveref} 
\usepackage{setspace}
\usepackage{booktabs}

\Crefformat{figure}{#2Fig.~#1#3}
\Crefmultiformat{figure}{Figs.~#2#1#3}{ and~#2#1#3}{, #2#1#3}{ and~#2#1#3}
\Crefformat{equation}{#2Eq.~#1#3}

\usepackage{xspace}
\makeatletter
\DeclareRobustCommand\onedot{\futurelet\@let@token\@onedot}
\def\@onedot{\ifx\@let@token.\else.\null\fi\xspace}
\makeatother

\newcommand{\p}{\mathbf{p}}

\newcommand{\x}{\mathbf{x}}
\newcommand{\m}{\mathbf{m}}
\newcommand{\obs}{\boldsymbol{\rho}}

\newcommand{\g}{\mathbf{g}}

\newcommand{\tc}{\mathbf{v}}
\newcommand{\cmd}{\mathbf{c}}
\newcommand{\action}{\mathbf{u}}

\definecolor{forestgreen}{RGB}{34,139,34}
\definecolor{orange}{RGB}{255, 191, 0}

\definecolor{lightblue}{HTML}{9BB7D4} 
\definecolor{light2blue}{HTML}{92A8D1} 
\definecolor{aqua}{HTML}{7BC4C4} 
\definecolor{turquoise}{HTML}{53B0AE} 
\definecolor{iris}{HTML}{5A5B9F} 

\definecolor{darkyellow}{HTML}{F0C05A} 
\definecolor{yellow}{HTML}{F8D948} 

\definecolor{pink}{HTML}{D94F70} 
\definecolor{lightpink}{HTML}{F7CAC9} 
\definecolor{warmpink}{HTML}{FF6F61} 
\definecolor{darkpink}{HTML}{C74375} 
\definecolor{magenta}{HTML}{BB2649} 

\definecolor{orange}{HTML}{E2583E} 
\definecolor{lightorange}{HTML}{FEBE98} 
\definecolor{orangered}{HTML}{DD4124} 

\definecolor{green}{HTML}{009473} 
\definecolor{grassgreen}{HTML}{88B04B} 

\definecolor{redbrown}{HTML}{955251} 
\definecolor{darkred}{HTML}{9B1B30} 

\definecolor{purple}{HTML}{6968AC} 
\definecolor{darkpurple}{HTML}{5F4B8B} 
\definecolor{lightpurple}{HTML}{B163A3} 

\definecolor{grey}{HTML}{959A9C} 
\definecolor{beige}{HTML}{DECDBE} 

\hypersetup{
    colorlinks=true, 
    linkcolor=black, 
    citecolor=darkred, 
    filecolor=blue, 
    urlcolor=grey,
}

\title{\LARGE \bf Teaching Robots Like Dogs: Learning Agile Navigation from Luring, Gesture, and Speech}

\author{Taerim~Yoon$^{1}$,
        Dongho~Kang$^{2}$,
        Jin~Cheng$^{2}$,
        Fatemeh~Zargarbashi$^{2}$,
        Yijiang~Huang$^{2}$, \\
        Minsung~Ahn$^{3}$,
        Stelian~Coros$^{2}$, and 
        Sungjoon~Choi$^{1}$
\thanks{\emph{Corresponding author: Sungjoon Choi.}}
\thanks{$^{1}$Taerim~Yoon, and Sungjoon~Choi are with the Department of Artificial Intelligence, Korea University, 145 Anam-ro, Seongbuk-gu, Seoul, Korea (email: taerimyoon@korea.ac.kr; sungjoon-choi@korea.ac.kr).}
\thanks{$^{2}$Dongho~Kang, Jin~Cheng, Fatemeh~Zargarbashi, Yijiang~Huang, and Stelian Coros are with the Department of Computer Science, ETH Zurich, Wasserwerkstrasse 12, 8092 Zurich, Switzerland (email: kangd@ethz.ch, jicheng@ethz.ch, fzargarbashi@ethz.ch, huang@hey.com, scoros@ethz.ch).}
\thanks{$^{3}$Minsung~Ahn is with the Department of Mechanical and Aerospace Engineering, UCLA, 420 Westwood Plaza, Los Angeles, 90095, CA, USA (email: aminsung@ucla.edu)}
}

\begin{document}
\maketitle
\pagestyle{plain}


\begin{abstract}
In this work, we aim to enable legged robots to learn how to interpret human social cues and produce appropriate behaviors through physical human guidance.
However, learning through physical engagement can place a heavy burden on users when the process requires large amounts of human-provided data.
To address this, we propose a human-in-the-loop framework that enables robots to acquire navigational behaviors in a data-efficient manner and to be controlled via multimodal natural human inputs, specifically gestural and verbal commands.
We reconstruct interaction scenes using a physics-based simulation and aggregate data to mitigate distributional shifts arising from limited demonstration data. 
Our \emph{progressive goal cueing} strategy adaptively feeds appropriate commands and navigation goals during training, leading to more accurate navigation and stronger alignment between human input and robot behavior. 
We evaluate our framework across six real-world agile navigation scenarios, including jumping over or avoiding obstacles.
Our experimental results show that our proposed method succeeds in almost all trials across these scenarios, achieving a 97.15\% task success rate with less than 1 hour of demonstration data in total.
\end{abstract}

\section{Introduction}
Dogs have remarkable abilities to interpret human behavior through social cues and seamlessly perform given tasks.
For instance, they can follow a pointing gesture, track a handler’s gaze, or respond to brief verbal commands with precise actions.
In contrast, current robot operation still primarily relies on manual control interfaces such as joysticks or keyboards.
These methods require continuous attention and fine-grained control, limiting their practicality for everyday use by non-experts.
Motivated by this, we aim to develop a system that maps natural human cues to corresponding robot behaviors.

Building such a system involves two core challenges.
First, single-modal commands, such as verbal or gestural instructions alone, introduce ambiguity.
For instance, verbal commands such as “go there” are inherently ambiguous unless the system correctly interprets the user’s pointing gesture that specifies the target location.
Second, collecting large-scale real-world human–robot interaction data to learn such multimodal associations is costly.
Therefore, it is crucial that the system can learn new interaction patterns from only a few interactions.


Prior work on natural robot interfaces has explored gesture-based control or language-conditioned navigation, but often relies on a single modality and substantial data collection.
In this paper, we introduce a human-in-the-loop framework for teaching and controlling quadruped robots through multimodal interactions, termed \textbf{LURE} (\textbf{L}uring-based \textbf{U}ser–\textbf{R}obot \textbf{E}ngagement), which enables data-efficient learning from few demonstrations.
Inspired by the \emph{luring} technique used by professional dog trainers, users first physically guide the robot with a teaching rod while issuing gestures and verbal commands. 
We then reconstruct the interaction scenes in a physics-based simulator, where the robot learns navigation skills that overcome distributional shifts arising from limited data, i.e., mismatches between the training and deployment distributions \citep{ross2011reduction}.
During training, we adaptively determine intermediate commands and navigation goals to improve context-aligned navigation, a process we refer to as \emph{progressive goal cueing}.
At run time, the learned policy enables intuitive control via human gestures and verbal commands, allowing the robot to perform agile movements in cluttered environments.

\begin{figure}
    \centering
    \includegraphics[width=0.92\linewidth]{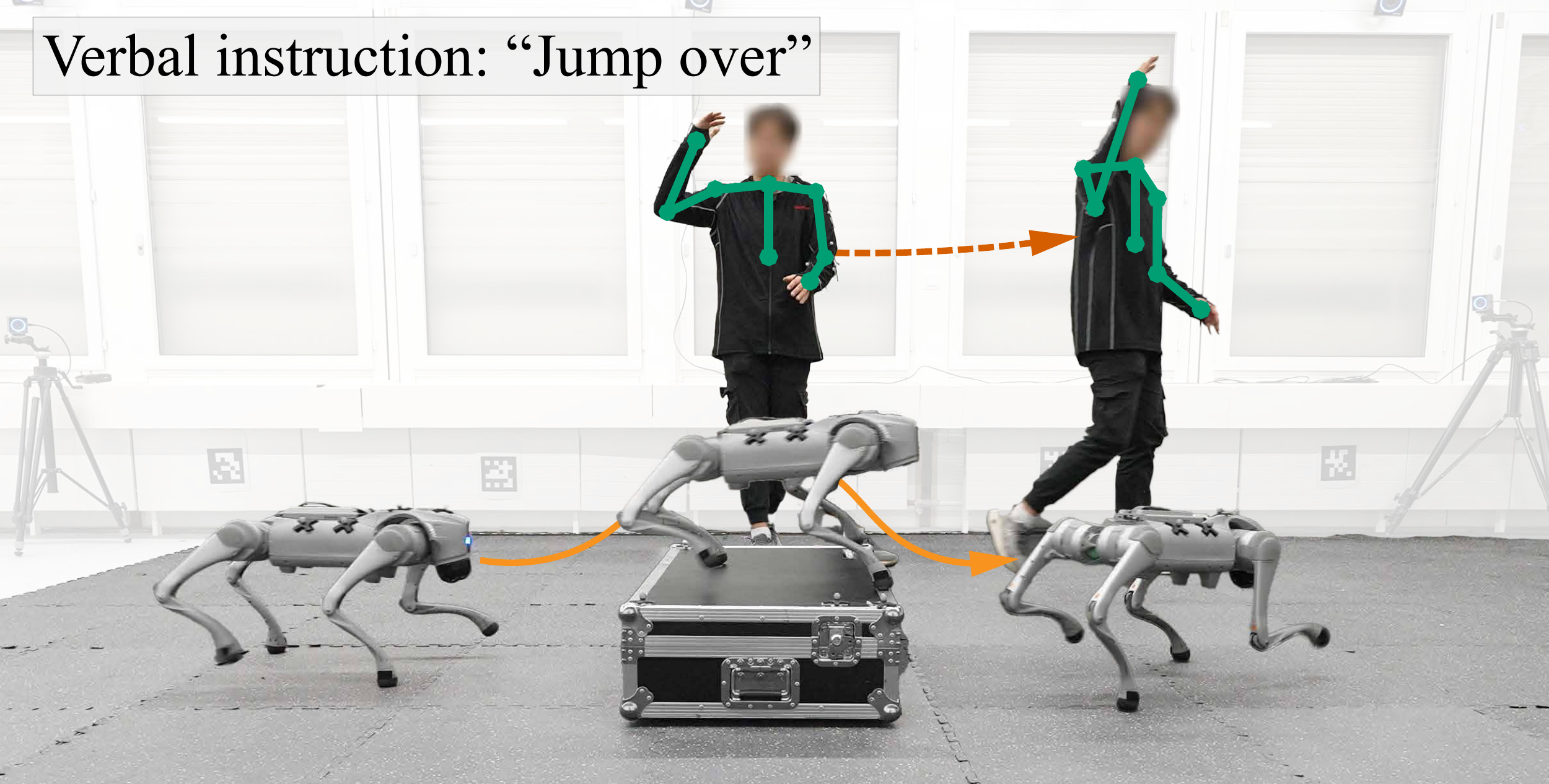}
    \caption{The proposed framework enables agile robot behaviors through verbal and gesture-based commands.}
    \label{fig:teaser_jumpover}
\end{figure}

We demonstrate that the proposed framework enables users to control quadruped robots to overcome obstacles using verbal and gesture commands.
For instance, in the \emph{Jump over} scenario shown in \Cref{fig:teaser_jumpover}, the robot correctly interprets the user’s intention to jump over the box and executes the desired action. 
Extensive experiments show that our system enables navigation through cluttered environments in a data-efficient manner with less than 1 hour of data across six scenarios.

To summarize the key contributions of this work: 
\begin{enumerate}
    \item We present a system that enables users to teach and control quadruped robots in cluttered environments through real-time gestural and verbal commands.
    \item We propose a data-aggregation technique that enables robots to learn new interaction patterns using only a few demonstrations.
    \item We experimentally show that verbal and nonverbal cues complement each other in interpreting user intent.
\end{enumerate}

\section{Related Work}
Our goal is to enable users to teach and control a legged robot through natural interactions.
This objective aligns with the human-in-the-loop control paradigm, as the robot is operated through human feedback.
Our approach also relates to learning from demonstrations, as we leverage interactive human demonstrations to acquire navigation skills.
In the following sections, we discuss related work on both aspects.

\subsection{Human-in-the-Loop Robot Interfaces}

Recent efforts in robot control have increasingly focused on designing more intuitive interfaces that move beyond manual devices such as joysticks or keypads.
In particular, controlling robots through human gestures and verbal commands has been a long-standing objective in human-robot interaction~\citep{goodrich2007human}.
While numerous studies have investigated the utility of these modalities, they have primarily focused on using either gestures or verbal commands in isolation.

A number of studies proposed a straightforward gesture-based human-robot interaction by manually mapping a predefined set of human gestures to a discrete number of robot actions.
For instance, \citet{shin2024non} and \citet{10553163} map current body keypoints directly to discretized robot actions.
While these approaches are simple, discrete actions lack expressiveness, motivating continuous gesture-to-action interfaces.
In this direction, \citet{xie2025human} transferred hand motions at the velocity level for continuous robot control, while \citet{cuan2024gesture2path} mapped user gestures to robot actions using imitation learning.
However, directly mapping hand motions to low-level robot actions requires continuous, exhaustive user attention, similar to joystick-based teleoperation.

Verbal command-based control interfaces have been explored, particularly in the context of language-conditioned navigation.
\citet{cheng2024navila} proposed a vision-language-action model for legged navigation, enabling text-driven goal specification.
More recently, \citet{xue2025leverb} embedded latent vision-language representations into a humanoid whole-body controller, enabling visual navigation from language instructions.
Despite these advances, many language-based systems still suffer from spatial ambiguities.
For instance, interpreting a vague command like “go there” can be challenging, as the destination is not explicitly specified in language but must be resolved through supplementary cues such as gaze or pointing.

To address these limitations, we complement the strength of verbal commands and gestures within a unified real-time interface.
By grounding language with accompanying user gestures, our system resolves spatial ambiguities and provides clearer intent signals for navigation.

\subsection{Learning Navigation from Demonstrations}
Learning from Demonstration (LfD) enables users to teach behaviors through examples, providing an intuitive way to teach by direct demonstration rather than relying on handcrafted rules~\citep{thomaz2006reinforcement, argall2009survey}.
In line with this, we adopt LfD as a natural framework for learning new behaviors from interactive and physically guided user demonstrations.

Early applications of LfD to navigation focused on single-goal imitation, with policies trained to follow expert-provided trajectories.
A pivotal work by \citet{pomerleau1989alvinn} trained neural networks to imitate human driving behavior for lane following, while more recent approaches \citep{wigness2022robot} guide path imitation using environment feature maps.

Recent work advances toward multi-skill navigation that interprets high-level commands and produces context-appropriate behaviors.
For example, \citet{wu2020towards} proposed a generative imitation learning framework for target-driven visual navigation in indoor scenes, enabling policies to generalize to novel goals based on demonstration data.
\citet{singh2022illiterate} proposed a goal-conditioned navigation framework that grounds implicit free-form language expressions into spatial targets, enabling mobile agents to interpret diverse instructions such as “go to the bed”.
Despite these advances, LfD is prone to distributional shift due to limited demonstration coverage, which can degrade performance or lead to task failure.

As in prior work, we study demonstration-based navigation, where robots navigate using verbal instructions and gestures.
In particular, we focus on mitigating distributional shifts under limited demonstrations via data augmentation.

\begin{figure}[!t]
    \centering
    \includegraphics[width=0.98\linewidth]{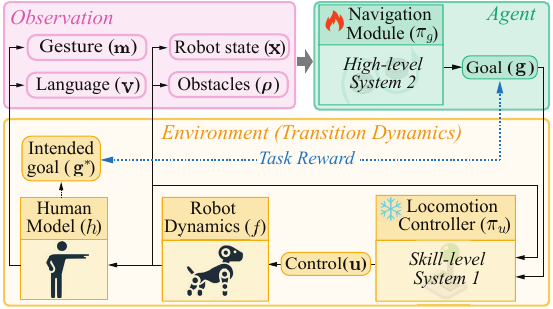}
    \caption{
    The problem is formulated as an MDP with observations $(\mathbf{x}, \boldsymbol{\rho}, \cmd)$ and actions representing the navigation goal $\mathbf{g}$.
    The transition dynamics include the pretrained locomotion controller $\pi_u$, the robot dynamics $f$, and the human model $h$, while the reward is defined as the alignment between the predicted and intended goals.
    }
    \label{fig:MDP}
\end{figure}

\section{Problem Formulation}
Let the robot state be $\x$, and the control input be $\action$, where $\x$ consists of the robot’s base position and orientation, joint angles, and their corresponding velocities, and $\action$ denotes the motor torque commands.
We assume that the robot dynamics are defined by a function $f$, such that $\x' = f(\x, \action)$, where $\x'$ denotes the state at the next time step.
The robot operates in a cluttered environment, where the surrounding environment is represented by a heightmap $\obs$.
Our objective is to enable a human user to control the robot to perform a navigation task $\mathcal{T}$ via interaction commands $\cmd$, consisting of a gesture command $\m$ and a verbal command $\tc$.
Specifically, we model the human as a function $h$ that produces an intended navigation goal $\g^*$ together with interaction commands $\cmd$, expressed as $(\cmd, \g^*) = h(\x, \obs, \mathcal{T})$.
Given the interaction commands $\cmd$, the robot interprets the user’s intent and moves toward the intended goal $\g^*$.
Formally, we develop a policy $\pi$ that generates control inputs $\mathbf{u} = \pi(\x, \cmd, \obs)$, enabling the robot to infer human intent and accomplish the navigation task $\mathcal{T}$ by reaching the intended goal $\g^*$.

As shown in \Cref{fig:MDP}, we decompose the problem into two subsystems: a \emph{high-level navigation module} and a \emph{low-level locomotion controller}.
The navigation module, denoted as $\pi_g$, serves as a high-level System~2 planner that interprets human intent and predicts navigation goals $\g$, represented as $\g = \pi_g(\cmd, \x, \obs)$.
The locomotion controller, represented as $\action = \pi_u(\g, \x, \obs)$, functions as a fast and reactive System~1 module that executes agile behaviors to overcome obstacles.
By integrating these two modules, the robot is controlled directly by high-level interaction commands.

We pretrain a locomotion controller $\pi_u$ capable of tracking arbitrary goals across diverse terrains and obstacles, and subsequently focus on the high-level navigation problem.
We formulate this problem as a Markov Decision Process (MDP) problem, where the observation is defined as $(\x, \obs, \cmd)$ and the action corresponds to the navigation goal $\g$ expressed in the robot frame.
In more detail, the navigation policy $\pi_g$ outputs navigation goal $\g$ based on the interaction signal $\cmd$, where the reward function is defined by how closely $\g$ matches the intended navigation goal $\g^*$.
The locomotion controller $\pi_u$ then takes the navigation goal $\g$ to produce a control signal $\action = \pi_u(\g, \x, \obs)$.
Subsequently, the robot dynamics $f$ produce the robot state $\x'$ at the next time step, and the human model $h$ generates the corresponding gesture $\m'$ and language command $\tc'$, completing the state transition.

We address this MDP using data aggregation~\citep{ross2011reduction} to mitigate the distributional shift problem, which is augmenting expert actions based on the current policy's state distribution, denoted as $\mathcal{D}^{\pi_g}$.
Formally, the problem is formulated as
\begin{equation} \label{eq:navigation_problem}
    \arg \min_{\pi_g \in \Pi} \, \mathbb{E}_{(\cmd, \x, \obs, {\g^*}) \sim \mathcal{D}^{\pi_g}} \left[ \left\| {{\g^*}} - \pi_g(\cmd, \x, \obs) \right\|^2 \right].
\end{equation}
Addressing distributional shifts is particularly important, as the high cost of collecting interaction data leads to limited demonstrations, making the system especially prone to them.

Data aggregation requires two key components.
First, it requires the transition dynamics, including the human model $h$. 
Since manually encoding the intended navigation goal $\g$ and the corresponding interaction commands $\cmd$ is impractical, we use demonstration data to approximate the human model as $\tilde{h}$, as discussed in \Cref{sec:progressive goal cueing}.
Second, data aggregation requires an expert policy to provide actions at previously unseen states. 
We construct a locally approximated expert $\tilde{\pi}_g$ that produces actions for novel states and enables efficient data aggregation, as described in \Cref{sec: data aggregation}.

\begin{figure*}[!t]
    \centering
    \includegraphics[width=0.98\linewidth]{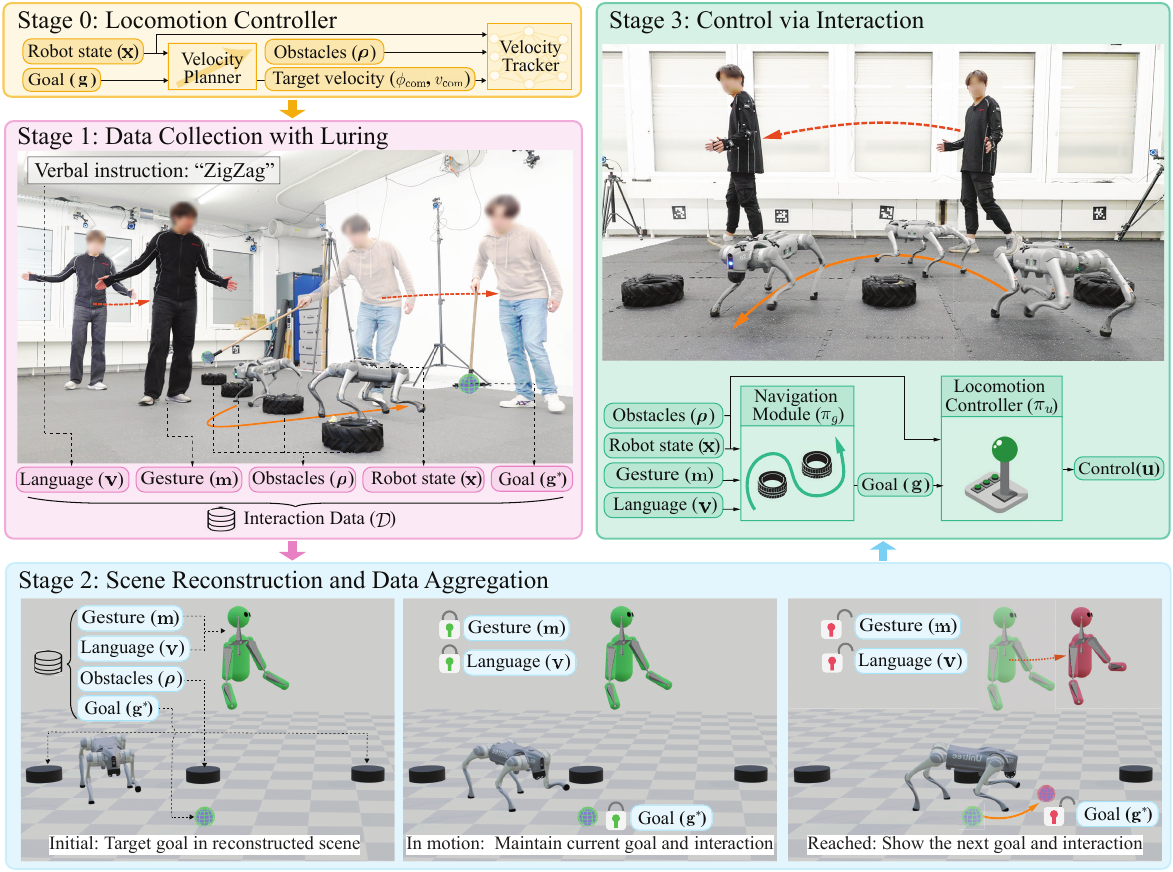}
    \caption{
    Overview of the proposed framework. We first establish a locomotion controller that follows the navigation goal $\g$. In Stage~1, we collect interaction data $\mathcal{D}=\{\tc, \m, \obs, \x, \g^*\}$ through natural interactions between two participants. In Stage~2, we reconstruct the interaction scene from $\mathcal{D}$ and train the navigation module via data aggregation. The framework progressively provides the interaction and navigation goals only after the robot reaches its current goal, ensuring alignment between the command and behavior. Finally, in Stage~3, the user can control the robot through interaction.
    }
    \label{fig:overview}
\end{figure*}


\section{Method}
Our goal is to develop a policy that can interpret the verbal command $\tc$ and gesture command $\m$ to reach the navigation goal $\g^*$.
To achieve this, our framework comprises several key components, as illustrated in \Cref{fig:overview}.
We begin by introducing an agile locomotion controller that enables the robot to perform dynamic movements such as stepping over or around obstacles.
Next, we collect interaction data using this locomotion controller.
Finally, we reconstruct interaction scenes in which robots are trained via data augmentation, enabling control through natural gestural and verbal commands with limited demonstration data.

\subsection{Locomotion controller}
As a preliminary stage, we develop a locomotion controller for low-level agility of the robot.
As shown in the Stage~0 from \Cref{fig:overview}, the locomotion controller consists of a velocity planner and a velocity tracker. 
The velocity planner converts the navigation goal $\g$ into a velocity command guiding the robot toward its target.
In more detail, it outputs a heading velocity command $v_\text{com} = K\|\g^{xy} - \p^{xy}\|_2$ and a rotation command $\phi_\text{com} = \text{yaw}(R^\top (\g - \p))$, where $\p \in \mathbb{R}^3$ is the robot's base position, $R \in \mathrm{SO}(3)$ is the robot's rotation matrix and $(\cdot)^{xy}$ indicates ground-projected coordinates.

The other component is the velocity tracker, which enables the robot to track the commanded velocity.  
Although our framework is compatible with any velocity tracker, we trained a specialized tracker designed for dynamic locomotion to support agile behaviors, such as obstacle jumping.
More details on our velocity tracking policy are discussed in \Cref{sec: details}.

\subsection{Human-in-the-loop data collection through luring}
We collect human–robot interaction data by luring, as illustrated in Stage~1 of \Cref{fig:overview}. 
The process requires two participants: one issues both verbal and non-verbal commands, while the other guides the robot using a teaching rod. 
This setup allows us to collect paired interaction and navigation data, which we refer to as the interaction data $\mathcal{D}$.
Details on the representation of verbal commands and gestural cues are provided in \Cref{sec: details}.

As shown in \Cref{fig:overview}, the interaction data $\mathcal{D}$ consists of the robot state $\x$, the target position $\g^*$, the gesture command $\m$, the verbal command $\tc$, and the obstacle representation $\obs$.
All data is recorded in a global coordinate frame, providing sufficient information for full scene reconstruction. 
This scene reconstruction is a crucial component of our framework, as it enables data aggregation that improves robustness to distributional shifts, as discussed in \Cref{sec: data aggregation}.

\subsection{Data aggregation} \label{sec: data aggregation}
We reconstruct the interaction scene in the simulation for data aggregation, as shown in Stage~2 of \Cref{fig:overview}.
Specifically, we recreate the terrain using obstacle information $\rho$ and replay the gesture $\m$ and verbal commands $\tc$.
The robot is trained to follow the intended goal location $\g^*$, defined by the teaching rod position in the interaction data.

Overall, our training procedure resembles the data aggregation method introduced in DAgger~\citep{ross2011reduction}, where the agent explores using its current policy while the expert’s actions are augmented to guide learning.
In more detail, we define the local expert policy $\tilde{\pi}_g$ that outputs the expert navigation goal $\tilde{\g}$ in the robot frame, directing the robot towards the global target position $\g^*$.
By aggregating these actions, the system can efficiently generate corrective actions in the local frame that steer the robot toward the global target $\g^*$, even when the current policy drifts away from the expert trajectory. 
As a result, the robot learns to recover from deviations and reach the intended target, effectively mitigating the distributional shift. 
This approach is justified by our data-collection setup, where the velocity planner drives the robot toward the goal in the robot frame, producing demonstrations consistent with the behavior encoded in the local expert policy $\tilde{\pi}_g$.

We apply domain randomization to deliberately force the agent to deviate from the expert trajectory, exposing it to a broader range of states and enabling it to learn recovery behaviors~\citep{kang2023rl}.
For example, we vary the robot’s mass and inertia, and apply external pushes to the base.
To further improve robustness, we introduce the following techniques, with detailed parameter settings provided in the appendix.
\begin{itemize}
    \item \textbf{Terrain scaling}
    We divide the terrain into a uniform grid of $S$ tiles and scale each tile by a random factor to randomize the spacing between obstacles and goals.
    \item \textbf{Binary obstacle map}
    We threshold the height map at $h_\text{thres}$ and feed the resulting binary obstacle map to the navigation module, thereby promoting consistent navigation across varying obstacle heights.
    \item \textbf{Distractor objects.}
    We introduce square and circular distractor objects to mitigate overfitting to specific obstacle arrangements.
\end{itemize}

\begin{figure*}[!t]
    \centering
    \captionsetup[subfloat]{labelfont=normal, font=normal}
    \subfloat[{Go there}]{
        \includegraphics[width=0.32\linewidth]{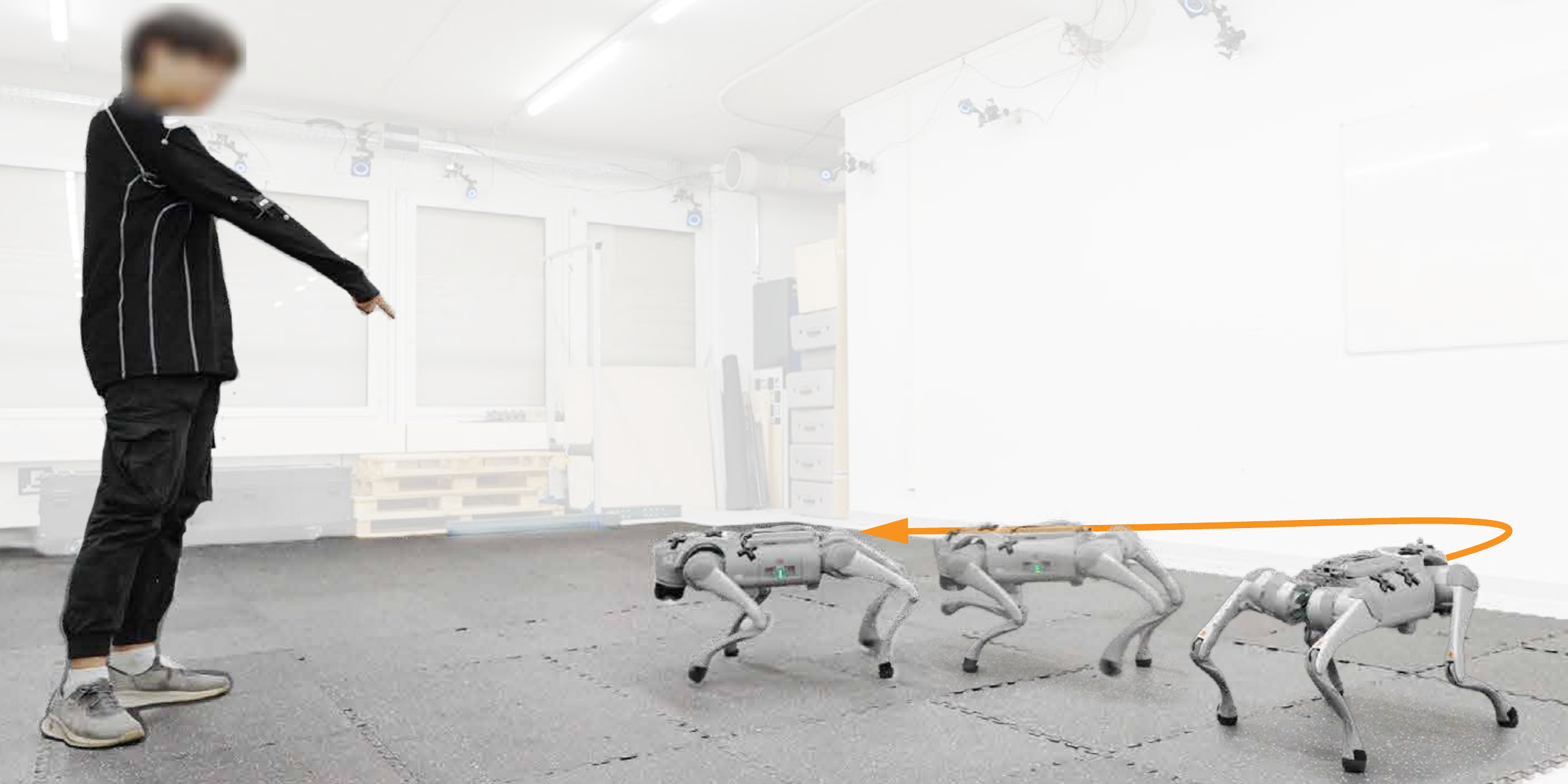}
        \label{fig:go there demo}
    }
    \subfloat[{Come here}]{\includegraphics[width=0.32\linewidth]{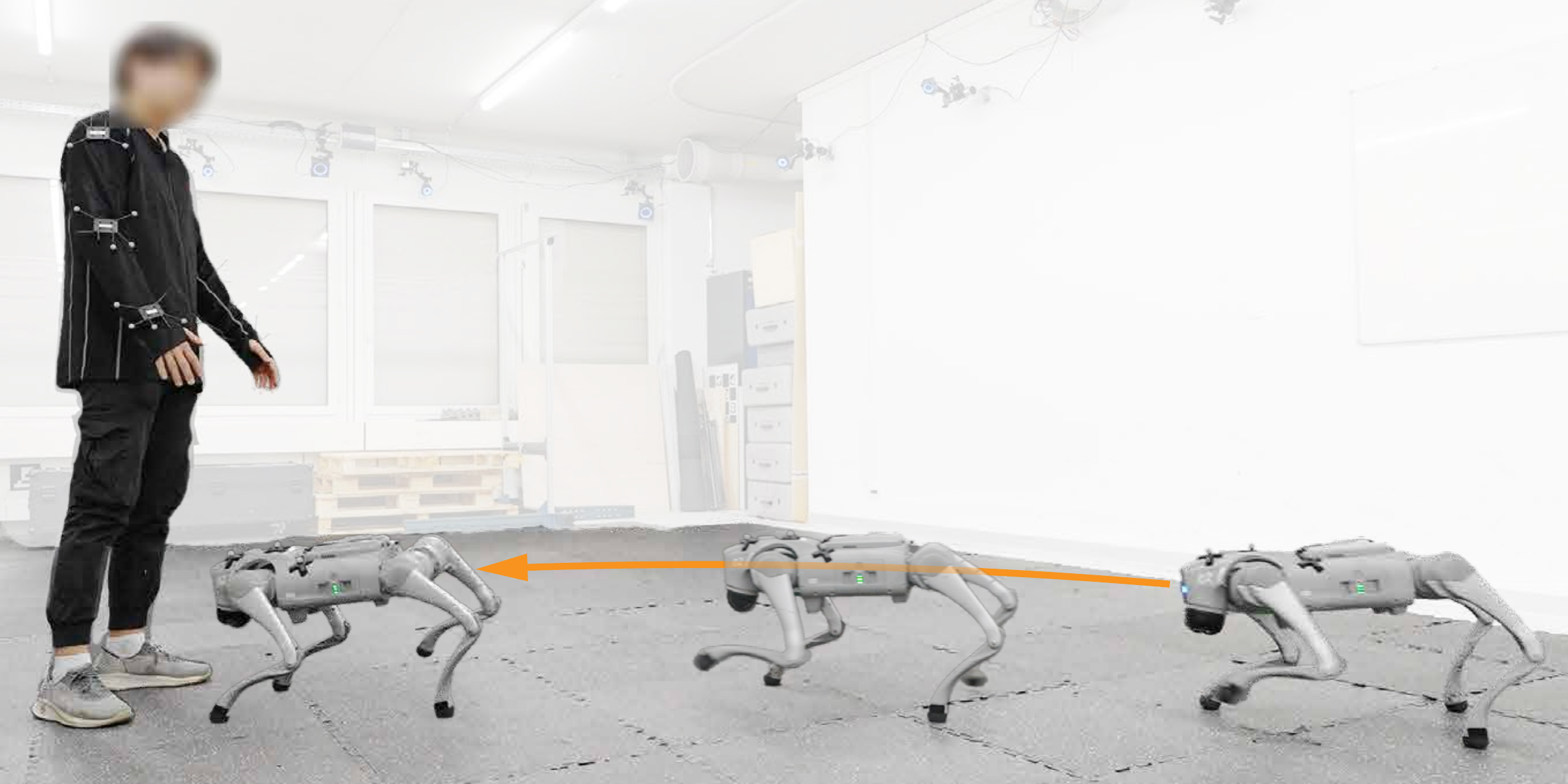}
        \label{fig:come here demo}
    }
    \subfloat[{Follow me}]{
        \includegraphics[width=0.32\linewidth]{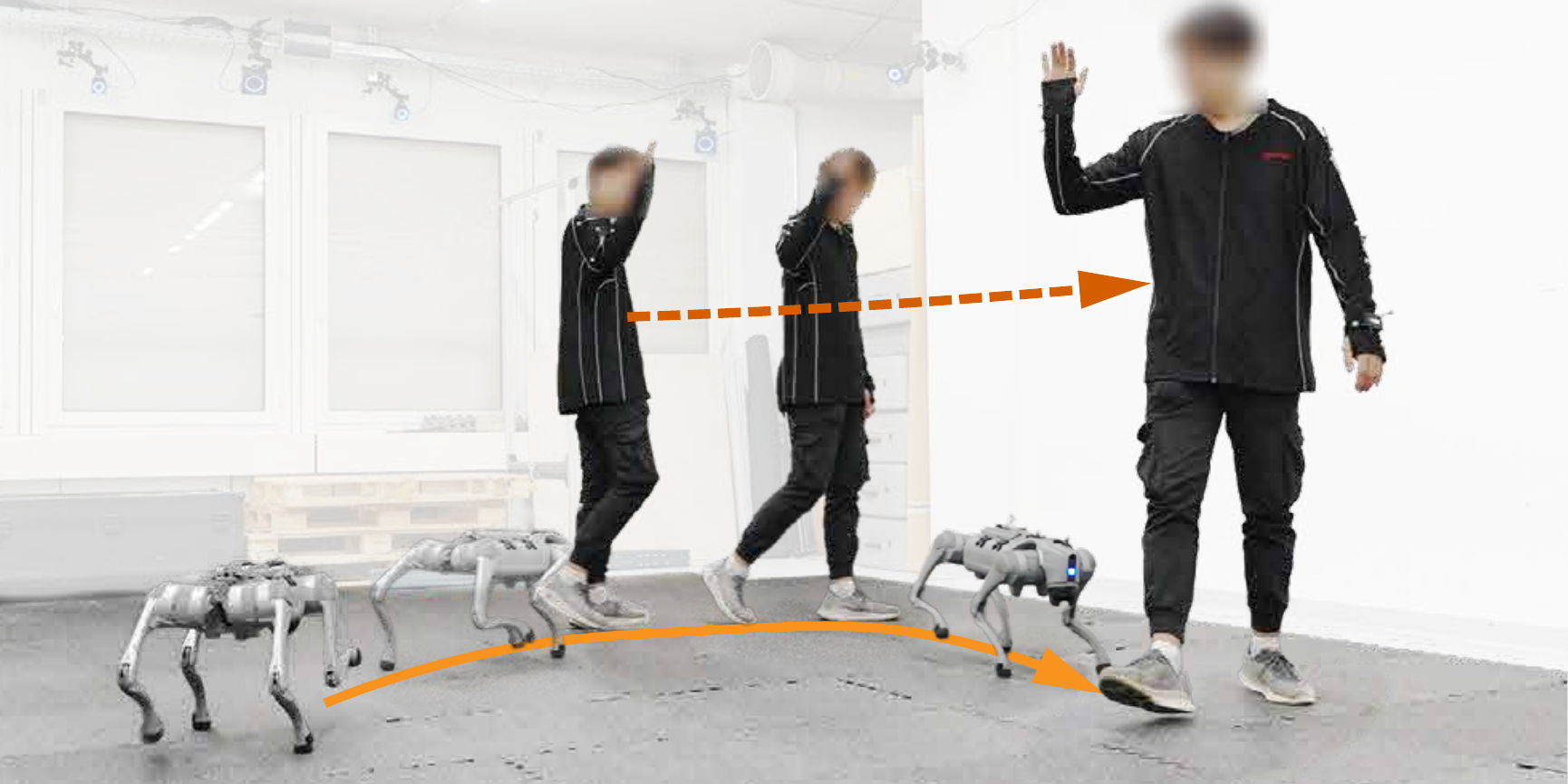}
        \label{fig:follow me demo}
    } \\
    \subfloat[{Come around}]{
        \includegraphics[width=0.32\linewidth]{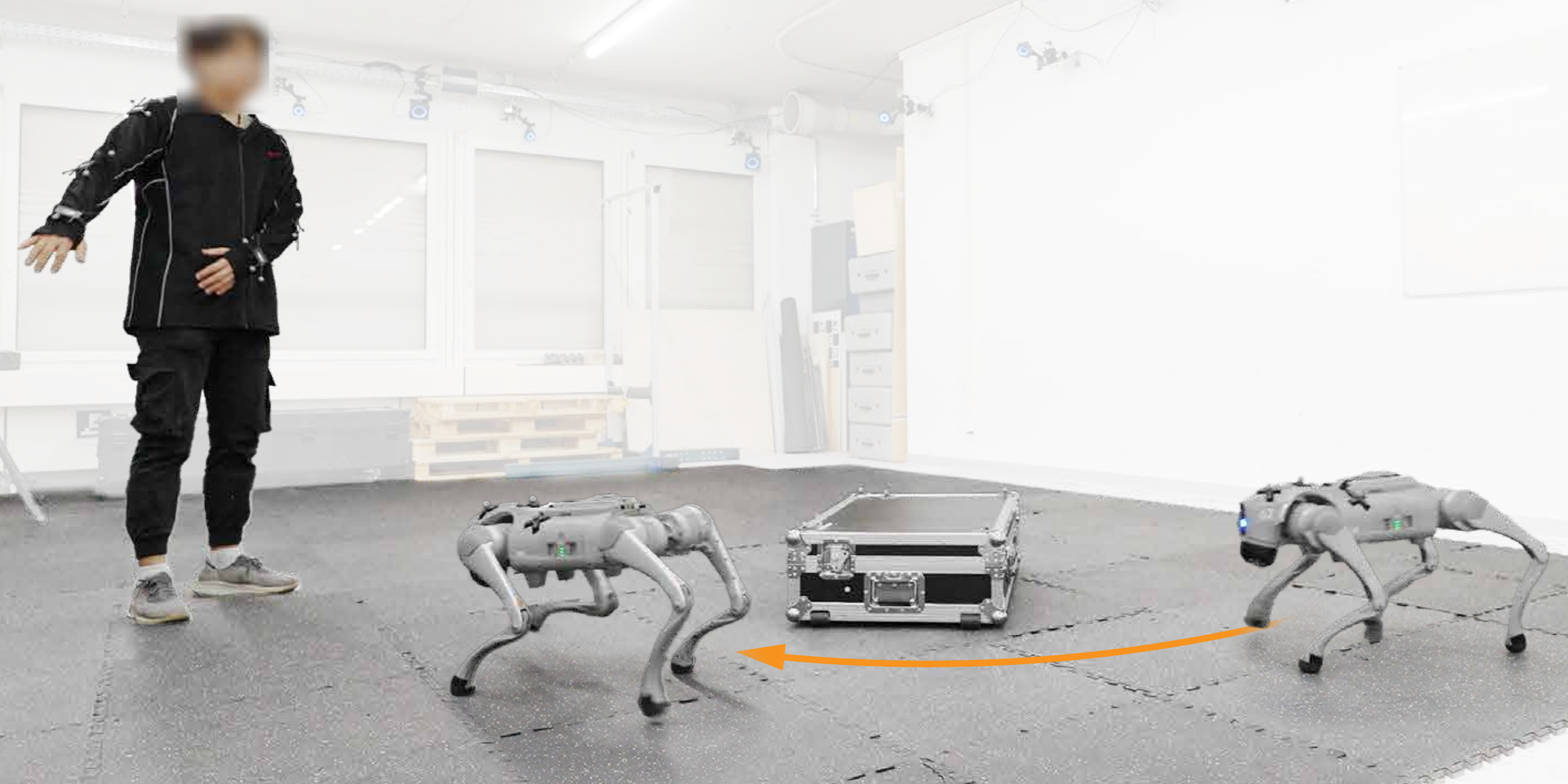}
        \label{fig:come around demo}
    } 
    \subfloat[{Jump over}]{
        \includegraphics[width=0.32\linewidth]{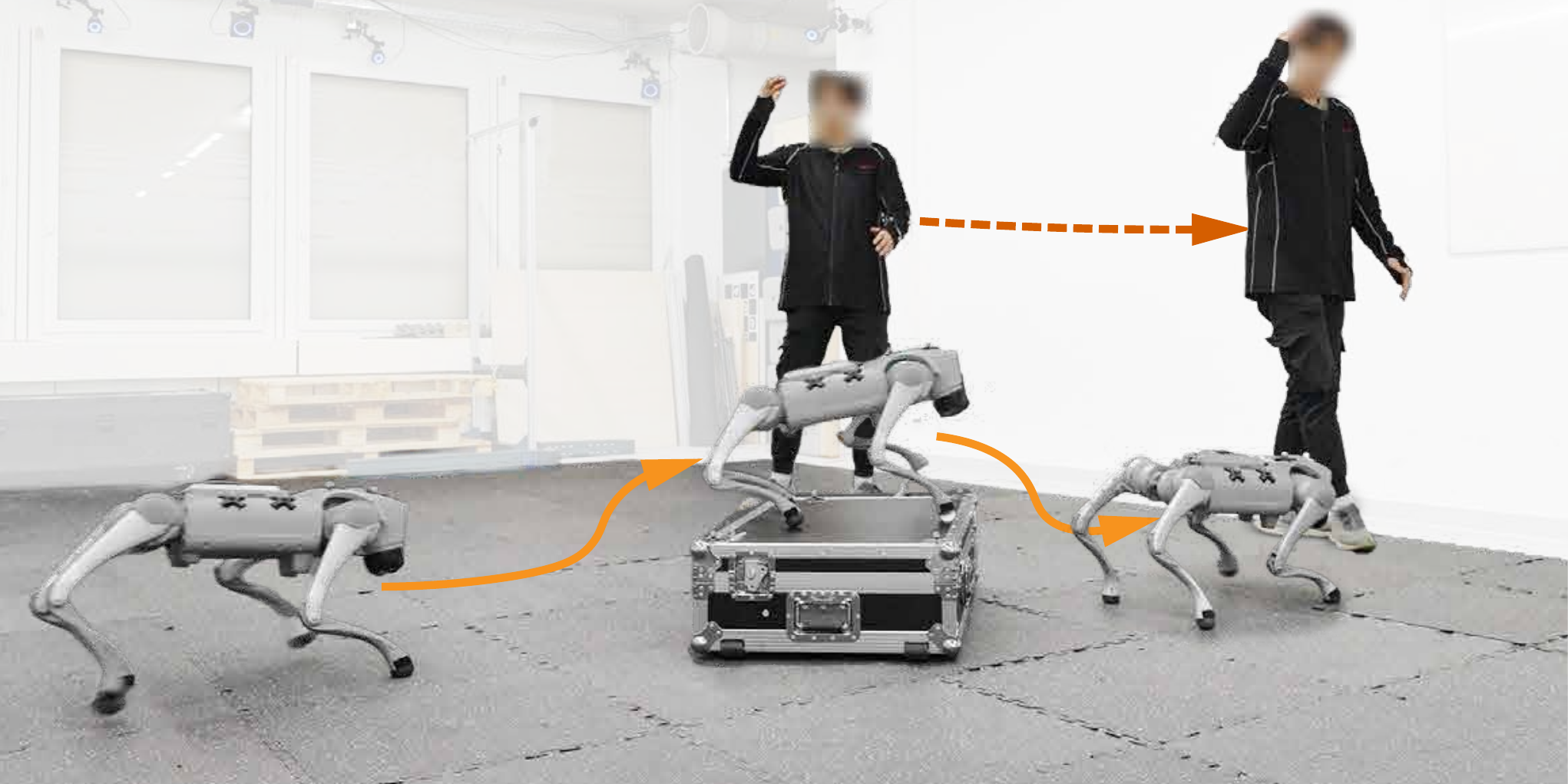}
        \label{fig:jump over demo}
    }
    \subfloat[{Zigzag}]{
        \includegraphics[width=0.32\linewidth]{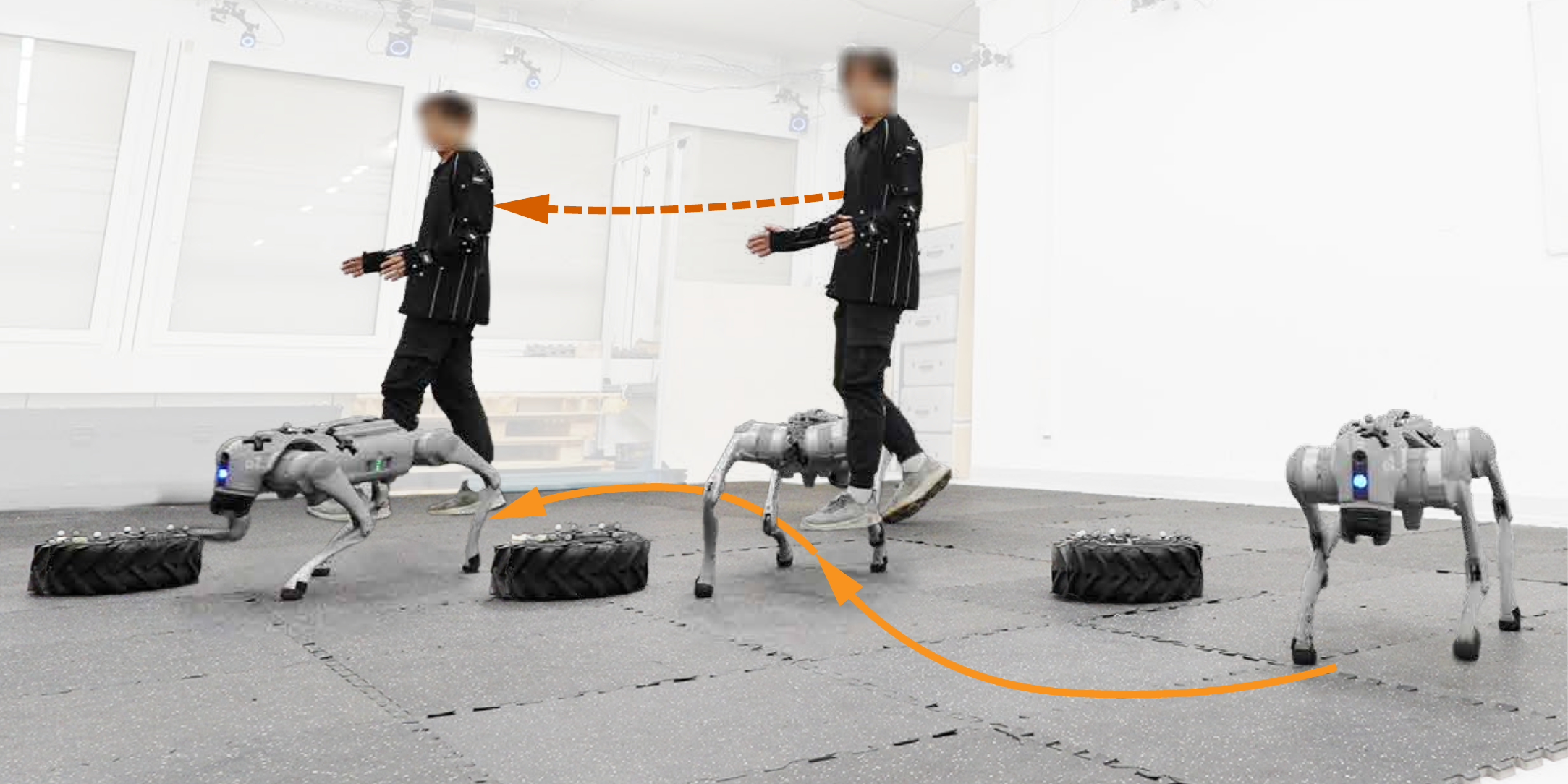}
        \label{fig:zigzag tire demo}
    }
    \caption{
        Illustration of data collection procedures for six interactive navigation scenarios. 
        (a) \textit{Go there}, (b) \textit{Come here}, and (c) \textit{Follow me} represent human-robot interactions in open space.
        (d) \textit{Come around} and (e) \textit{Jump over} illustrate interactions involving a box obstacle. 
        (f) \textit{Zigzag} demonstrates interaction with multiple tire obstacles.
    }
    \label{fig:scenarios}
\end{figure*}

\subsection{Progressive goal cueing} \label{sec:progressive goal cueing}
It is crucial to provide interaction commands and navigation goals that align with the robot’s current state during training, as misalignment can lead to inconsistent supervision.
One naive approach is to feed gesture and verbal commands to the policy according to their original timestamps.
However, during early training, the policy often progresses more slowly through tasks, leading to new commands being issued before the robot has completed the preceding behavior.
Since the interaction data were collected by issuing new commands in response to the robot’s task progress, this naive replay strategy can introduce state–command misalignment.

To address this, we propose a \emph{progressive goal cueing} that dynamically presents the interaction command based on the robot's state.
As illustrated in Stage~2 of \Cref{fig:overview}, each interaction command is held constant while the robot moves toward the current goal and is updated once the target is reached.
This allows the human model $h$ to wait for the robot rather than rushing through the episode when the robot struggles to navigate correctly in the early stages. 
Overall, it results in more context-aware control and enables finer movements during execution, as demonstrated later in \Cref{sec:SR}.

At each timestep, we apply progressive goal cueing with a 50\% probability; otherwise, it advances to the next interaction according to the simulation clock.
This design choice prevents excessive waiting and improves responsiveness during real-world deployment.
Additionally, the waiting behavior naturally fades out as the robot begins producing correct movements in the latter stage of training.

\subsection{Interaction Data Augmentation}
Since our model is trained from limited interaction data, preventing overfitting and promoting generalization across both gesture and verbal modalities are essential.
Therefore, we apply modality-specific input augmentation for both gestures and verbal commands.
For gestures, we perturb the extracted keypoints with noise $\epsilon_{\m}$ for every simulation time step.
This models the natural variability in repeated human motions as well as noise introduced during pose estimation.
For verbal commands, we leverage a Large Language Model (LLM)~\citep{achiam2023gpt} to generate semantically equivalent expressions.
Specifically, for each verbal command collected in \Cref{sec: data collection}, we prompt the LLM to produce $N_v$ alternative phrasings, from which one is randomly sampled during training.

\subsection{Implementation details}~\label{sec: details}
This section provides additional implementation details regarding the perception and command representations, as well as the velocity tracker policy for low-level locomotion control. 

\begin{figure}[!t]
    \centering
    \captionsetup[subfloat]{labelfont=normal, font=normal}
    \subfloat[{Data collection time}]{  \includegraphics[width=0.45\linewidth]{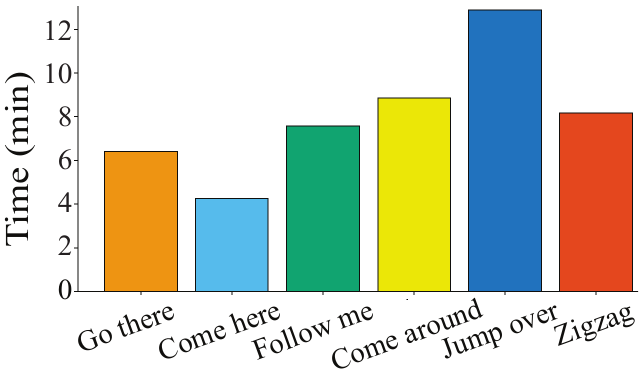}
        \label{fig:time portion}
    } 
    \subfloat[{Number of episodes}]{
        \includegraphics[width=0.45\linewidth]{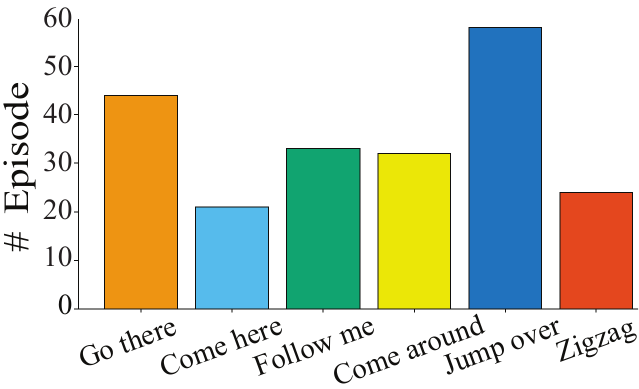}
        \label{fig:interaction number}
    }
    \caption{
        Analysis of collected data in terms of
        (a) total wall clock time and 
        (b) number of episodes for each scenario
    }
    \label{fig:data analysis}
\end{figure}

\subsubsection{Representation}~\label{sec: representation}
We describe how each input modality is represented and captured in real time.
The verbal command is segmented using Silero VAD~\citep{bredin2021end}, transcribed using Whisper~\citep{radford2023robust}, and encoded with a pretrained text encoder~\citep{wang2022text}. 
Next, the human gesture is captured using motion capture cameras (OptiTrack Prime X22). We consider six upper-body keypoints—the shoulders, elbows, and wrists on both the left and right sides—and express their 3D positions in the robot frame as $\mathbf{p}_{\text{sh.r}}, \mathbf{p}_{\text{el.r}}, \mathbf{p}_{\text{wr.r}}, \mathbf{p}_{\text{sh.l}}, \mathbf{p}_{\text{el.l}}, \mathbf{p}_{\text{wr.l}}$. Using these keypoints, we define the gesture command as
$\m = \big[ \mathbf{d}_{\text{sh.r,el.r}}, \ 
             \mathbf{d}_{\text{el.r,wr.r}}, \ 
             \mathbf{d}_{\text{sh.l,el.l}}, \ 
             \mathbf{d}_{\text{el.l,wr.l}}, \ 
             \mathbf{d}_{\text{sh.l,sh.r}}, \ 
             (\mathbf{p}_{\text{sh.l}} + \mathbf{p}_{\text{sh.r}})^{xy}/2, \ 
             \phi_h \big]$, 
where $\mathbf{d}$ denotes the unit vector between two points, and $\phi_h$ represents the human yaw.
Regarding the obstacles, we measure their dimensions in advance and estimate their poses using motion capture cameras.
Using this information, we obtain a heightmap in the robot frame for the locomotion controller $\pi_u$ and a thresholded binary mask for the navigation module $\pi_g$.

\subsubsection{Velocity tracker} \label{sec: velocity tracker}
We train a velocity tracker operating at 50~Hz to overcome the obstacles.
The training scheme is similar to the work of \citet{cheng2024extreme} and is trained in Isaac Gym~\citep{makoviychuk2021isaac} to track velocity commands $v_\text{com} \in [0.0, 1.0]~\mathrm{m/s}$ and angular velocity commands $w_\text{com} \in [-\pi/3, \pi/3]~\mathrm{rad/s}$.
To improve robustness, we introduce a stand boolean in the state representation by thresholding velocity commands $v_\text{com}$, which enhances walk–stand transitions.
For real-world deployment, we use a Kalman filter–based state estimator~\citep{bloesch2013state} that fuses joint encoder, foot contact, and motion capture data for stable feedback.
Although the robot can be trained to move faster, we limit its speed for safety.

\section{Results}
We conduct a series of experiments to assess the effectiveness and generality of our framework.
We begin by describing the interaction scenarios considered in this study and then compare our method with imitation learning baselines to evaluate our proposed data aggregation and progressive goal cueing strategy.
We then show that our system can quickly learn new interaction patterns from novel users with less than five minutes of interaction.
Next, we conduct an ablation study to demonstrate how verbal and gestural cues complement each other.
Finally, we demonstrate multi-obstacle navigation in a real-world environment, showing that our system can robustly execute complex task sequences using a single unified policy.

\subsection{Data Collection} \label{sec: data collection}
We collect interaction data, denoted as $\mathcal{D}$, comprising human interaction $\mathbf{\cmd}$, robot states $\mathbf{x}$, obstacle representations $\obs$, and target goals $\mathbf{g}^*$, recorded at a frequency of 10 Hz.
We define six interaction scenarios, as shown in \Cref{fig:scenarios}, and a stop scenario.
\begin{itemize}
    \item \textbf{Stop:} A person raises both arms and says ``stop,'' prompting the robot to halt at its current position.
    
    \item \textbf{Go there:} A person points with one hand and says ``go there,'' and the robot moves toward the indicated location.
    
    \item \textbf{Come here:} A person calls the robot by saying ``come here'' with the arms dropped, and the robot moves directly toward the person.
    
    \item \textbf{Follow me:} A person raises either their left or right hand and says ``follow me,'' and the robot follows on the corresponding side of the raised hand.
    
    \item \textbf{Come around:} A person stands in front of a box and says ``come around,'' prompting the robot to move toward the human while avoiding the obstacle. Both hands move towards the direction intended for coming around the obstacle (left or right).
    
    \item \textbf{Jump over:} A person waves one hand over their head while saying ``jump over,'' causing the robot to jump over the obstacle.
    \item \textbf{Zigzag:} A person gestures with one hand to indicate the movement direction while saying ``zigzag,'' and the robot navigates between the tires in a zigzag pattern when viewed from above.
\end{itemize}

The summary of total data collection time and the number of episodes per scenario is detailed in \Cref{fig:data analysis}.
While we introduce these scenarios as representative cases, the proposed framework can support a much broader range of interactions \emph{without relying on predefined cases}.

\subsection{Baselines} \label{sec: baselines}
We compare our proposed method to three imitation learning algorithms.

\subsubsection{BC~\citep{pomerleau1989alvinn}}
As a most basic baseline, we train the navigation policy using standard Behavior Cloning (BC). 
The policy is learned purely from expert demonstrations.

\subsubsection{GAIL~\citep{ho2016generative}}
We use Generative Adversarial Imitation Learning (GAIL) as a baseline to compare against inverse reinforcement learning methods that are known for their generalization capabilities.
The policy is learned via adversarial training to match the expert behavior distribution.
This baseline allows us to compare our approach against an imitation framework that focuses on generalization.

\subsubsection{DAgger~\citep{ross2011reduction}}~\label{sec: dagger}
We train the policy using data aggregation with a local expert $\tilde{\pi}_g$ consistent with our proposed framework.
However, we do not apply progressive goal cueing. 
Instead, interaction commands are issued according to the simulation time clock.
We reconstruct the scene and apply domain randomization to induce a distributional shift, from which the policy learns robust recovery, as described in \ref{sec: data aggregation}.

\subsubsection{LURE (Ours)}
Our proposed method, LURE, utilizes data aggregation and domain randomization as in DAgger. In addition, we apply progressive goal cueing.

\begin{figure*}[!t]
    \centering
    \captionsetup[subfloat]{labelfont=normal, font=normal}
    \subfloat[Success Rate]{
        \includegraphics[width=0.47\linewidth]{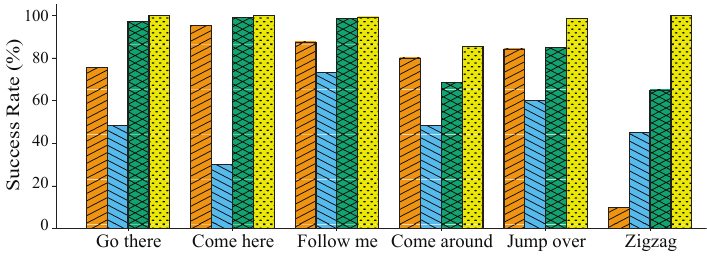}
        \label{fig:success_rate_plot}
    }
    \subfloat[Navigation Error]{
        \includegraphics[width=0.47\linewidth]{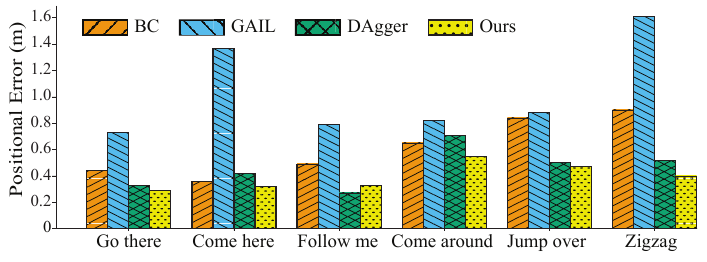}
        \label{fig:pos_error_plot}
    }
    \caption{
        Comparison of baseline methods across six interactive navigation scenarios.
        (a) Navigation error (m) of BC, GAIL, DAgger, and our proposed method.
        (b) Success rate (\%) for the same scenarios.
    }
    \label{fig:baseline_results}
\end{figure*}

\begin{table*}[!t]
\centering
\small
\caption{Success Rate and Navigation Error per Scenario, Subject, and Adaptation}
\setlength{\tabcolsep}{4.0pt}
\begin{tabular}{lcccccccccccc}
\toprule
\multirow{3}{*}{\textbf{Scenario}} 
& \multicolumn{4}{c}{\textbf{Subject \#1}} 
& \multicolumn{4}{c}{\textbf{Subject \#2}} 
& \multicolumn{4}{c}{\textbf{Subject \#3}} \\
\cmidrule(lr){2-5} \cmidrule(lr){6-9} \cmidrule(lr){10-13}
& \multicolumn{2}{c}{Success Rate $\uparrow$} & \multicolumn{2}{c}{Navigation Error $\downarrow$}
& \multicolumn{2}{c}{Success Rate $\uparrow$} & \multicolumn{2}{c}{Navigation Error $\downarrow$}
& \multicolumn{2}{c}{Success Rate $\uparrow$} & \multicolumn{2}{c}{Navigation Error $\downarrow$} \\
\cmidrule(lr){2-3} \cmidrule(lr){4-5}
\cmidrule(lr){6-7} \cmidrule(lr){8-9}
\cmidrule(lr){10-11} \cmidrule(lr){12-13}
& w/o Adpt. & Adpt. & w/o Adpt. & Adpt.
& w/o Adpt. & Adpt. & w/o Adpt. & Adpt.
& w/o Adpt. & Adpt. & w/o Adpt. & Adpt. \\
\midrule
Go there
& 0.66 & \textbf{0.87} & 0.71 & \textbf{0.53}
& 0.86 & \textbf{0.99} & 0.57 & \textbf{0.51}
& 0.68 & \textbf{1.00} & 0.72 & \textbf{0.54} \\
Come here
& 0.90 & \textbf{1.00} & 0.39 & \textbf{0.28}
& \textbf{1.00} & \textbf{1.00} & 0.38 & \textbf{0.18}
& \textbf{1.00} & \textbf{1.00} & 0.26 & \textbf{0.22} \\
Follow me
& 0.88 & \textbf{1.00} & 0.68 & \textbf{0.36}
& 0.81 & \textbf{0.90} & 0.50 & \textbf{0.30}
& \textbf{1.00} & \textbf{1.00} & 0.39 & \textbf{0.23} \\
Come around
& 0.37 & \textbf{0.95} & 1.13 & \textbf{0.41}
& \textbf{0.95} & \textbf{0.95} & 0.65 & \textbf{0.43}
& 0.10 & \textbf{1.00} & 2.20 & \textbf{0.89} \\
Jump over
& 0.87 & \textbf{0.95} & 1.08 & \textbf{0.53}
& 0.73 & \textbf{0.95} & 0.70 & \textbf{0.55}
& 0.48 & \textbf{0.92} & 1.11 & \textbf{0.45} \\
Zigzag
& 0.76 & \textbf{0.95} & 0.99 & \textbf{0.36}
& 0.50 & \textbf{0.70} & 1.45 & \textbf{0.73}
& 0.60 & \textbf{1.00} & 1.04 & \textbf{0.36} \\
\bottomrule
\end{tabular}
\label{tab:subject_adaptation}
\end{table*}

\subsection{Evaluating Success Rate and Navigation Error} \label{sec:SR}
We measure both the robot's success rate and navigation error, as shown in \Cref{fig:baseline_results}. 
We first segment each interaction scene based on the issued verbal command “Stop,” treating each segment as an independent episode.
For example, the period between saying “Go there” and issuing “Stop” is regarded as one episode. 
At the beginning of each episode, we reset the robot's position to the ground-truth position from data to ensure consistent evaluation. 
During evaluation, we disable progressive goal cueing to avoid underestimating the performance of other baselines.
For each interaction episode, we evaluate performance on five terrain instances with different scale factors and 20 robots per terrain, resulting in 100 evaluation runs.
The reported success rates and navigation errors are then averaged over each scenario.

\paragraph{Task Success}  
We define task-specific success criteria as follows:
\begin{itemize}
    \item \textbf{Go there, Come here, Follow me:} The final position of the robot must be within 1 meter of the ground-truth base position.
    \item \textbf{Come around:} The robot must pass through the correct lateral corridor beside the box, following the direction indicated by the user.
    \item \textbf{Jump over:} The robot must pass through three reference points located along the forward direction of the box, centered at $-0.5$ m, $0$ m, and $+0.5$ m relative to the box. A pass is counted as successful if the robot comes within 0.2 m of each point.
    \item \textbf{Zigzag:} The robot must pass through all midpoints between the tires while avoiding stepping on any of them. A pass is counted as successful if the robot comes within 0.2 m of each point.
\end{itemize}

\paragraph{Navigation Error}  
We compute the navigation error as the Mean Squared Error (MSE) between the robot’s base position and a ground-truth trajectory, measured at each timestep during the interaction.
The ground-truth trajectory corresponds to the robot motion recorded in the interaction data under the same interaction commands.
The final error is obtained by averaging the per-frame MSE over the entire interaction duration. 
This metric reflects how accurately the policy follows the intended trajectory specified by the interaction commands.

The results summarized in \Cref{fig:baseline_results} show that LURE achieves the best performance, with a clear margin, attaining an average success rate of 97.15\%.
By comparing DAgger with BC, we observe that incorporating data aggregation reduces navigation error by 24.6\%. 
Introducing progressive goal cueing, on top of this, yields an additional 15.2\% reduction in navigation error over the DAgger baseline. 
In terms of success rate, data aggregation improves performance by 18.6\%, and progressive goal cueing provides a further 13.7\% gain.
Notably, GAIL outperforms BC on the most challenging Zigzag task, but its overall performance degrades when all skills are considered together.
On average, GAIL achieves a 29.32\% lower success rate than BC, which may be explained by training instability and mode collapse often observed in adversarial imitation learning.



\begin{table}[!t]
\centering
\small
\caption{Success Rate (\%) and Navigation Error (m) per Scenario and Modality}
\begin{tabular}{lcccccc}
\toprule
\multirow{2}{*}{\textbf{Scenario}} &
\multicolumn{3}{c}{\textbf{Succ. Rate $\uparrow$}} &
\multicolumn{3}{c}{\textbf{Pos. Error (m) $\downarrow$}} \\
\cmidrule(lr){2-4} \cmidrule(lr){5-7}
 & Both & Verb. & Gest. & Both & Verb. & Gest. \\
\midrule
Go there     & \textbf{0.85} & 0.35 & 0.15 & \textbf{0.52} & 1.15 & 1.79 \\
Come here    & \textbf{0.91} & 0.01 & 0.15 & \textbf{0.40} & 1.74 & 1.09 \\
Follow me    & \textbf{0.98} & 0.16 & 0.28 & \textbf{0.29} & 1.98 & 1.24 \\
Come around  & \textbf{0.90} & 0.02 & 0.22 & \textbf{0.54} & 2.26 & 1.85 \\
Jump over    & \textbf{0.83} & 0.67 & 0.23 & \textbf{0.49} & 1.03 & 1.64 \\
Zigzag       & \textbf{0.96} & 0.00 & 0.16 & \textbf{0.35} & 1.93 & 1.26 \\
\bottomrule
\end{tabular}
\label{tab:modality_combined}
\end{table}

\subsection{Adaptation to novel user} \label{sec: adaptation}
We evaluate whether our system can generalize to novel users through quick adaptation. 
This is essential because different users often employ different interaction cues to command the robot for the same task.
We first train a model using the initial interaction dataset from a single user described in \Cref{sec: data collection}, and then fine-tune it using an average of 4.5 minutes of data collected from each new subject.
For evaluation, we measure the success rate and navigation error in the reconstructed simulation environment, following the same protocol described in \Cref{sec:SR}.

As shown in \Cref{tab:subject_adaptation}, the success rates before adaptation are 74.00\%, 80.83\%, and 64.33\% for three subjects.
This indicates that the model works to some degree even without personalization. 
After adaptation, the success rates increase to 95.33\%, 91.50\%, and 98.67\%, resulting in an average improvement of 22.11\%.
These results demonstrate that our framework can quickly adapt to new users with only a small amount of data.

\begin{figure}
    \centering
    \includegraphics[width=0.95\linewidth]{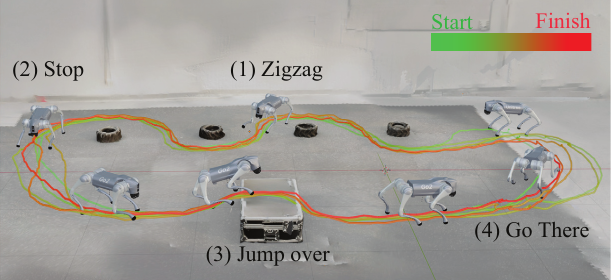}
    \caption{
    The robot (1) zigzags through the tires, (2) briefly stops, (3) jumps over the box, and (4) returns to the starting point according to the real-time user's instruction.
    The visualization overlays the robot’s trajectory on a reconstructed 3D scene of the real environment.
    The figure shows the robot completing these sequences five consecutive times.
    }
    \label{fig:course run}
\end{figure}

\subsection{Ablation on Gesture and Verbal Modalities} \label{sec: ablation}

We use both gesture and verbal modalities for interaction cues.
To evaluate their individual contributions, we train a model in which each modality is masked with a 10\% probability during training.
This enables the model to function even when one modality is absent, enabling meaningful evaluation under partial inputs.

As shown in \Cref{tab:modality_combined}, the evaluation is conducted across all six scenarios to measure how each modality contributes to task success and navigation error. 
Using gesture only results in an average success rate of 19.83\%, which is 70.67\% lower than when both modalities are used. 
In practice, the robot also tends to move only when a gesture is held for an extended period. 
These indicate that gesture alone is often insufficient for effectively interpreting the user’s intent.
On the other hand, using verbal commands alone results in a navigation error of 1.6817 $\mathrm{m}$, whereas combining both modalities reduces the error by 1.25 $\mathrm{m}$ to 0.4317 $\mathrm{m}$.
This is because verbal commands can lack spatial grounding.
In the \emph{Go there} scenario, for instance, the robot does not know where the user is pointing and begins to move randomly. 
One exception is the \emph{Jump over} scenario, where the robot can locate the box without a gesture, yielding a relatively high success rate of 67\%. However, this is still lower than the multimodal setting, which achieves 83\%.

\subsection{Multi-Obstacle Navigation} \label{sec: course running}
To demonstrate that our framework can robustly handle diverse obstacles while seamlessly switching between multiple locomotion skills based on user intent, we conducted an obstacle course experiment, as illustrated in \Cref{fig:course run}.
The course contains both boxes and tires: the robot first navigates through the tire corridor, avoiding each tire according to the user’s interaction commands.
After clearing the tires, the user issues a “stop” command, upon which the robot halts and then performs the “Jump over” command to overcome the box.
Finally, in response to the “Go there” command, the robot returns to the starting point.
As shown in \Cref{fig:course run}, our framework completes this entire sequence five times in a row, demonstrating strong real-world robustness and smooth task transitions under a single unified policy.
The robot’s trajectory is obtained using a motion capture system, while the obstacle geometries are reconstructed using a consumer-grade mobile 3D scanning software.

\section{Limitations}
While our framework enables effective interaction-based learning and control, it currently lacks zero-shot generalization across different users due to substantial inter-subject variability in gesture styles.
Unlike verbal commands, which benefit from semantic augmentation via Large Language Models (LLMs), our gesture representation is augmented only through noise injection and therefore lacks semantic diversity.
A promising direction is to develop gesture representations that are semantically structured and jointly grounded in the robot state and surrounding objects, building upon pretrained motion representations such as \citet{zhang2024motiondiffuse}.

Moreover, our progressive goal cueing provides a simple and efficient way to approximate human behavior in novel states.
However, it only adjusts the timing of replaying human interaction data based on the robot’s state.
Ideally, human behavior should be generated more dynamically in response to novel states, rather than being constrained to temporally re-aligned demonstrations.

Finally, our current vision system is restricted to controlled environments, as the objects are modeled as circular or square primitives with pre-measured dimensions, and the human wears a motion capture suit for pose estimation. 
Camera-based human pose estimation and scene reconstruction~\citep{shin2024wham, mihajlovic2025volumetricsmpl} could potentially remove these assumptions and enable operation in the wild.

\section{Conclusion and Future Work}
We presented a framework that enables robots to learn and execute navigation skills in real time through natural human interaction. 
By reconstructing interaction scenes and augmenting data within these environments, our approach enables policies to robustly recover from distributional shifts across diverse interaction conditions. 
Experimental results demonstrate that the proposed framework effectively integrates verbal and gestural cues, enabling robust control across a range of tasks and environments. 
Moreover, our findings highlight the complementary nature of speech and gesture, showing that their combination provides a powerful interface for teaching and controlling robots to execute context-appropriate behaviors, including agile maneuvers such as jumping over obstacles. 
Overall, this work advances an intuitive and practical system that directly maps high-level human intent to executable robot actions. 
In future work, we will explore extending our framework beyond navigation to more general whole-body behaviors, such as loco-manipulation, by grounding multimodal commands using a similar data augmentation strategy.

\bibliographystyle{IEEEtranN}
\bibliography{root.bib}

\clearpage

\appendix
\begin{table}[h]
    \centering
    \caption{Hyperparameters and data augmentation details}
    \label{tab:appendix_hyperparams}
    \setlength{\tabcolsep}{6pt}
    \begin{tabular}{l p{0.65\columnwidth}}
        \toprule
        \textbf{Category} & \textbf{Setting} \\
        \midrule

        \makecell[l]{Domain\\Randomization}
        & \makecell[{{p{0.65\columnwidth}}}]{%
            External perturbation magnitude: $f \in [0, 0.5]~\mathrm{m/s}$; applied at an average interval of 3 seconds.\\
            Terrain scaling factor: $s \in [0.75, 1.25]$ over $3 \times 3$ tiles.\\
            Binary height map threshold: $0.05~\mathrm{m}$.} \\

        \addlinespace[0.5em]
        \hline
        \addlinespace[0.5em]

        \makecell[l]{Data\\Augmentation}
        & \makecell[{{p{0.65\columnwidth}}}]{%
            LLM-based verbal command augmentation: $N_v = 20$.\\
            Gesture keypoint noise: $\epsilon_{\m} \sim \mathcal{N}(0, 0.01)$.} \\
        \bottomrule
    \end{tabular}
\end{table}
\end{document}